\documentclass[10pt,twocolumn,letterpaper]{article}

\usepackage[pagenumbers]{template/iccv} %

\usepackage{multirow}
\usepackage{multicol}
\usepackage{amsmath}
\usepackage{algorithm}
\usepackage{algpseudocode}
\usepackage{dblfloatfix}
\usepackage{tabularx}
\usepackage{wrapfig}

\definecolor{iccvblue}{rgb}{0.21,0.49,0.74}
\usepackage[pagebackref,breaklinks,colorlinks,allcolors=iccvblue]{hyperref}

\def\paperID{10373} %
\def\confName{ICCV}
\def\confYear{2025}

\title{Controlling Multimodal LLMs via Reward-guided Decoding}

\author{First Author\\
Institution1\\
Institution1 address\\
{\tt\small firstauthor@i1.org}
\and
Second Author\\
Institution2\\
First line of institution2 address\\
{\tt\small secondauthor@i2.org}
}
\author{
Oscar~Mañas\textsuperscript{1,2,4}, ~Pierluca~D'Oro\textsuperscript{1,2,4}, ~Koustuv~Sinha\textsuperscript{4}, \\
Adriana~Romero-Soriano\textsuperscript{1,3,4,5}, ~Michal~Drozdzal\textsuperscript{4}, ~Aishwarya~Agrawal\textsuperscript{1,2,5} \\[0.5em]
\textsuperscript{1}Mila~-~Quebec~AI~Institute, ~\textsuperscript{2}Université~de~Montréal, ~\textsuperscript{3}McGill~University, \\
\textsuperscript{4}Meta~FAIR, ~\textsuperscript{5}Canada~CIFAR~AI~Chair \\[0.5em]
{\normalsize\tt oscar.manas@mila.quebec}
}

\begin{document}
\maketitle
\begin{abstract}
As Multimodal Large Language Models (MLLMs) gain widespread applicability, it is becoming increasingly desirable to adapt them for diverse user needs.
In this paper, we study the adaptation of MLLMs through controlled decoding. To achieve this, we introduce the first method for reward-guided decoding of MLLMs and demonstrate its application in improving their visual grounding.
Our method involves building reward models for visual grounding and using them to guide the MLLM's decoding process.
Concretely, we build two separate reward models to independently control the degree of object precision and recall in the model's output.
Our approach enables on-the-fly controllability of an MLLM's inference process in two ways: first, by giving control over the relative importance of each reward function during decoding, allowing a user to dynamically trade off object precision for recall in image captioning tasks; second, by giving control over the breadth of the search during decoding, allowing the user to control the trade-off between the amount of test-time compute and the degree of visual grounding.
We evaluate our method on standard object hallucination benchmarks, showing that it provides significant controllability over MLLM inference, while consistently outperforming existing hallucination mitigation methods.\looseness-1
\end{abstract}

\section{Introduction}
\label{sec:intro}

Multimodal Large Language Models (MLLMs) have shown great potential to solve a wide range of visiolinguistic tasks, while offering a general language interface to users~\cite{bordes2024introduction,caffagni2024r}.
As the adoption of MLLMs increases~\cite{achiam2023gpt,team2023gemini,dubey2024llama}, the demand to easily control their behavior to satisfy diverse user needs is emerging. Two needs, in particular, arise among the most important for users of MLLMs: 
a) control over the precision and thoroughness of their output (e.g., object recall), and b) control over the amount of compute spent to generate those outputs.
For instance, a user with visual impairment using the system to understand their surroundings may want the MLLM to respond with highly precise outputs (as hallucinations might be highly undesirable), while avoiding overly high latency on limited compute (e.g., on a smartphone); instead, a user leveraging the MLLM to generate synthetic captions to train downstream models may prioritize more diverse and detailed outputs (even if it means tolerating lower precision) while having the flexibility to spend more compute.

In this paper, we tackle this problem and propose a method for inference-time alignment of MLLMs.
Our method, called multimodal reward-guided decoding (MRGD), employs two reward functions, one tailored for hallucination reduction~\cite{bai2024hallucination} and one tailored for improving object recall.
Using these reward functions as criteria for searching for better outputs, our method gives control over the two axes mentioned above: by giving the option to set a relative weight for each reward,
it allows to smoothly control the trade-off between object precision and recall of the MLLM's outputs; by varying the breadth of the search, we can control the trade-off between the amount of test-time compute and the degree of visual grounding (which encompasses both object precision and recall).

Previous works explored reducing hallucinations in MLLMs by using methods such as prompting~\cite{zhangmultimodal}, supervised fine-tuning (SFT)~\cite{liumitigating} and RLHF fine-tuning~\cite{sun2023aligning,yu2024rlhf,zhou2024aligning}.
However, these methods do not allow fine-grained inference-time controllability of the MLLM's behavior: prompting relies on very coarse control by means of prompt engineering, while SFT and RLHF allow no controllability at all during inference.
For text-only LLMs, reward-guided decoding has been shown to be an effective way of obtaining fine-grained controllability~\citep{mudgalcontrolled,deng2023reward,khanovargs,snell2024scaling}, but there is a lack of such techniques for MLLMs.
Compared to the text-only case, for which a reward model processes data from a single modality, reward models guiding MLLMs face unique challenges, as they need to process both visual and textual information at the same time.
In particular, multimodal reward models need to understand the interaction between the generated text output and an input from a different modality (an image).
This interaction can cause specific types of hallucinations to emerge~\cite{zhouanalyzing} and should be addressed by tailored solutions~\citep{sun2023aligning}.\looseness-1

\begin{figure*}[ht]
    \vspace{-1em}
    \centering
    \includegraphics[width=\linewidth]{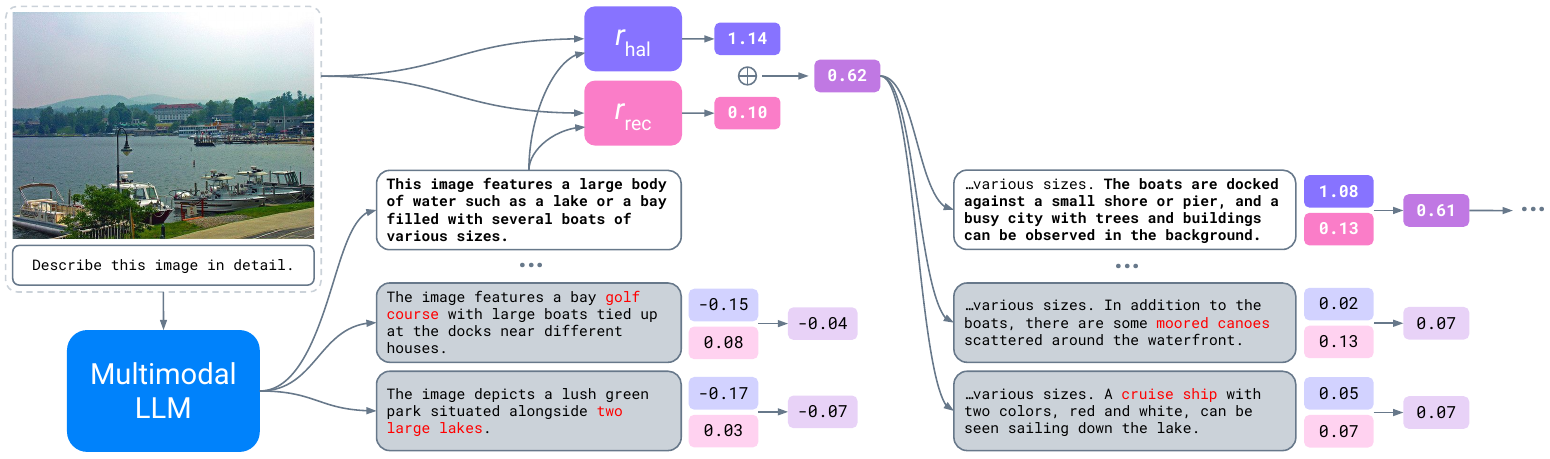}
    \caption{Illustration of multimodal reward-guided decoding (MRGD) for MLLMs. At each iteration, $k$ candidate completions (sentences in our case) to a partial response are sampled from the MLLM and evaluated according to a linear combination of rewards (the process is illustrated for the first selected completion and omitted elsewhere). The completion with largest score is selected and added to the context to generate the next $k$ candidates, until the \texttt{<EOS>} token is encountered.}
    \label{fig:mrgd_diagram}
    \vspace{-1em}
\end{figure*}

In summary, the main contributions of our paper are:\looseness-1
\begin{itemize}
    \item We propose a novel approach for reward-guided decoding for MLLMs, based on building reward models for different aspects of visual grounding and combining them to guide the search for high-quality outputs at test time.\looseness-1
    \item Through extensive experiments, we show that in MLLMs there exists an inherent trade-off between object precision and recall, as well as between compute and visual grounding quality. Our proposed method allows a user to specify a desired balance between these factors, enabling smooth adaptive control over the precision and recall, as well as between compute and visual grounding trade-offs, depending on task requirements and resource constraints.\looseness-1
    \item We demonstrate on standard hallucination benchmarks that our proposed method consistently outperforms existing hallucination mitigation approaches, while allowing test-time controllability of an MLLM's outputs.\looseness-1
\end{itemize}

\section{Related work}
\label{sec:related}

\paragraph{Guided decoding of LLMs.}
In the text-only setting, several works have explored guiding the decoding process of LLMs with a reward model to control output features such as helpfulness and harmlessness, and summary quality.
\cite{dathathriplug,khanovargs} train a reward model to evaluate full responses, and apply it at each decoding step to evaluate response prefixes and modulate the next-token probability distribution before sampling.
Instead, \cite{yang2021fudge,deng2023reward,mudgalcontrolled,han2024value,rashid2024critical} explicitly train a scoring function to predict the expected reward of response prefixes, also known as value function.
\cite{li2024cascade,brown2024large,snell2024scaling,lew2023sequential,liu2024don} explore sampling strategies such as best-of-$k$, beam search, or Monte Carlo tree search, which are based on generating multiple responses and selecting the best one with a reward model or value function.
Unlike existing methods for LLMs, we build \emph{multimodal} reward models to evaluate responses to \emph{multimodal} instructions, which additionally contain images, and focus on evaluating MLLMs on visual grounding tasks.
These models require processing both visual and textual information simultaneously, which can lead to different types of hallucinations that are specific to the multimodal nature of the inputs.
Therefore, it is important to develop solutions that effectively address these unique challenges.

\noindent\textbf{Mitigating hallucinations of MLLMs.}
Prior work on mitigating visual hallucinations of MLLMs has focused on supervised fine-tuning~\cite{liumitigating,liu2024investigating}, preference fine-tuning with RLHF/RLAIF ~\cite{sun2023aligning,zhao2023beyond,yu2024rlhf,zhou2024aligning,yu2024rlaif,zhou2024calibrated,amirloo2024understanding,sarkar2024mitigating,wang2024mdpo} or prompting~\cite{zhangmultimodal}.
Reward-guided decoding can be more powerful than fine-tuning or prompting, as it directly optimizes the output, making it more likely to produce the desired results~\citep{mudgalcontrolled,deng2023reward,khanovargs,snell2024scaling}. In contrast, the principles learned during fine-tuning or specified in (system) prompts may not always be respected at generation time.
In addition, reward-guided decoding can be combined with prompting or fine-tuning, and readily applied to many base models without retraining.
Other methods also propose to refine the MLLM's output, either via post-hoc rectification~\cite{zhouanalyzing,yin2023woodpecker,zhang2025self} or specialized decoding strategies~\cite{zhao2024mitigating,wan2024contrastive,favero2024multi,leng2024mitigating}.
Most similar to our method, CGD~\cite{deng2024seeing} uses an off-the-shelf CLIP-like model to guide decoding.
However, we show that training a multimodal reward model on preference data using a stronger backbone is more effective at mitigating hallucinations.
Furthermore, we propose to guide the decoding process with a combination of reward models for visual grounding, which allows the user to control the trade-off between object hallucination and recall in the MLLM's outputs.\looseness-1

\section{Method}
\label{sec:method}

We propose a multimodal reward-guided decoding strategy to improve the controllability of MLLMs at inference time. We first build small yet effective multimodal reward models to evaluate different aspects of visual grounding, and later combine them for search-based guided decoding.\looseness-1

\subsection{Building multimodal reward models}

The effectiveness of our guided decoding strategy hinges on the existence of a reward function capable of successfully evaluating how well a response satisfies a certain objective. Unlike for math or coding problems~\cite{brown2024large}, there are no automatic verifiers for the open-ended responses generated by MLLMs. 
We want to build a method that gives controllability over the outputs of an MLLM by trading off object precision and recall at inference time.
To achieve so, we build two reward models (RMs), that allow to incentivize precision and recall respectively: (1) an object hallucination reward model $r^{\theta}_{\text{hal}}$ (shortened to $r_\text{hal}$ when omitting the parameters is clear), trained from preference data obtained from a mixture of datasets, and (2) a recall reward model $r_\text{rec}$, obtained by combining pre-trained modules.
We next describe in detail how we build these two reward models.

\subsubsection{Training $r_\text{hal}$ from preference data}
\label{sec:paligemma_preference}

Given a dataset of multimodal preference data for object hallucinations $D = \{x_v, x_q, y^+, y^-\}_i$, where $y^+$ and $y^-$ are the chosen and rejected responses respectively, we train our reward model for object hallucination $r^{\theta}_{\text{hal}}$ as a classifier that predicts the preference probability following the Bradley-Terry model~\cite{bradley1952rank,ouyang2022training}:
\begin{equation}
\begin{aligned}
    \mathcal{L}_{RM}(x_v, x_q, y^+, y^-; \theta) = -\log\sigma(&r^\theta_{\text{hal}}(x_v, x_q, y^+) \\
    &- r^\theta_{\text{hal}}(x_v, x_q, y^-))
\end{aligned}
\end{equation}

To facilitate combining multiple rewards, it is desirable that $r^\theta_{\text{hal}}(x_v, x_q, y) \in [0,1]$. Therefore, we add a pair of mean-squared error loss terms to encourage $r^\theta_{\text{hal}}(x_v, x_q, y^+)$ to be close to $1$ and $r^\theta_{\text{hal}}(x_v, x_q, y^-)$ to be close to $0$, while simultaneously avoiding the gradient saturation pitfalls of squashing activation functions. 
Ultimately, this leads to the following loss function:
\begin{equation}
\begin{aligned}
    \mathcal{L}(\theta) = \mathop{\mathbb{E}}_{(x, y^+, y^-)\sim D}[&\mathcal{L}_{RM}(x, y^+, y^-; \theta) \\
    &+ (r^\theta_{\text{hal}}(x, y^+) - 1)^2 + r^\theta_{\text{hal}}(x, y^-)^2],
\end{aligned}    
\end{equation}
where $x = (x_v, x_q)$.

We use PaliGemma~\cite{beyer2024paligemma} as the backbone of our reward model for object hallucination, and add to it a regression head consisting of a linear layer projecting the last output token embedding to a single scalar. During our initial exploration, we also considered CLIP~\cite{radford2021learning} as a potential backbone for our reward model, but we ultimately discarded it due to the limited context length of CLIP's text encoder, which was insufficient to handle the longer responses present in preference data.

\subsubsection{Building $r_\text{rec}$ by composing off-the-shelf modules}

We build our reward model for object recall $r_\text{rec}$ from three off-the-shelf modules: a pre-trained object detector, a pre-trained word embedding model, and classical NLP tools. Given an image $x_v$ and a generated caption $y$, we first extract the reference objects from the image with the object detector, denoted as $O_{ref} = \{o_1, o_2, ..., o_n\}$, where $n$ is the number of detected reference objects. We also extract the predicted objects from the caption with a POS tagger, denoted as $O_{pred} = \{p_1, p_2, ..., p_m\}$, where $m$ is the number of generated objects.
We embed both reference and predicted objects with word embeddings $f_e: \mathcal{W} \rightarrow \mathbb{R}^d$, where $\mathcal{W}$ is the set of all words and $d$ is the dimensionality of the embedding space. This results in embedded reference objects $E_{ref} = \{f_e(o_1), f_e(o_2), ..., f_e(o_n)\}$ and embedded predicted objects $E_{pred} = \{f_e(p_1), f_e(p_2), ..., f_e(p_m)\}$.
Next, we compute the all-to-all semantic similarity between the embedded reference and predicted objects using a similarity function $\mathrm{sim}: \mathbb{R}^d \times \mathbb{R}^d \rightarrow \mathbb{R}$. Specifically, for each predicted object $p_i$, we compute its similarity with each reference object $o_j$ as $\mathrm{sim}_{ij} = f_e(p_i) \cdot f_e(o_j)^T$.
We consider a predicted object $p_i$ as a true positive if its maximum semantic similarity with any reference object is above a threshold $\tau$, i.e., $\max_{j=1,...,n} \mathrm{sim}_{ij} > \tau$.
Finally, we sum all true positives and divide by the number of reference objects to obtain the estimated object recall $r_\text{rec}$:\looseness-1
\begin{equation}
r_\text{rec} = \frac{\sum_{i=1}^m \mathbb{I}(\max_{j=1,...,n} \mathrm{sim}_{ij} > \tau)}{n},
\end{equation}
where $\mathbb{I}(\cdot)$ is the indicator function.

\subsection{Multimodal reward-guided decoding}

Our goal is to guide the generation process of an MLLM where the generated response is modulated using the two reward functions described above. 
Given an image $x_v$ and a visual instruction $x_q$, an MLLM $\pi$ generates a text response $y = \{y_1, ..., y_n\}$ autoregressively token-by-token, i.e., $y \sim \pi(x_v, x_q)$. 
To give a user the possibility of choosing the relative strength of each reward model
on-the-fly, we define a score $s$ as the linear combination of the rewards for object hallucination $r_\text{hal}$ and object recall $r_\text{rec}$:
\begin{equation}
\begin{aligned}
    s(x_v, x_q, y) &= w \cdot r_\text{hal}(x_v, x_q, y) \\
    &+ (1-w) \cdot r_\text{rec}(x_v, x_q, y),
    \label{eq:weighted_score}
\end{aligned}
\end{equation}
where $w \in [0,1]$ is a guidance strength hyperparameter chosen at inference time. A user can modulate the strength of the reward guidance by varying $w$.
At the extremes, for $w{=}1$, the best response is chosen entirely by following the reward model for object hallucination, while for $w{=}0$ only the reward model for object recall is applied.

\setlength{\textfloatsep}{1em}
\begin{algorithm}[t]
\caption{Multimodal reward-guided decoding}
\label{alg:mrgd}
\begin{algorithmic}
    \State $y \gets \texttt{""}$
    \While{$\texttt{<EOS>} \notin y$}
        \State $Y \gets \emptyset$
        \For{$j = 1$ to $k$}
            \State $y' \sim \pi(x_v, x_q, y)$
            \State $Y \gets Y \cup \{y'\}$
        \EndFor
        \State $y' = \arg\max_{y' \in Y} s(x_v, x_q, [y; y'])$
        \State $y \gets [y; y']$
    \EndWhile
\end{algorithmic}
\end{algorithm}

Given the score $s(x_v, x_1, y)$, we search for a response by expanding a search tree of partial responses and deciding which partial response to complete depending on the reward-based selection criterion.
At each iteration, we sample $k$ candidate completions $Y=\{ y^j_{i..i+m} \}_{j=1}^k$ from a single partial response, with $(m < n)$, evaluate each of them with a reward-based score $s(x_v,x_q,y^j_{1..i+m})$, select the one with the maximum score, and add it to the context. We then iterate this process until the \texttt{<EOS>} token is generated (see Algorithm~\ref{alg:mrgd}).\looseness-1

Since a reward model's score for a partial response also depends on how well-formed the text is, evaluating a partial response at an arbitrary token can produce a lower score for a partial response that may be in reality more aligned than others (due to, e.g., incomplete words).
To address this potential issue, we take advantage of the fact that captions are typically composed of multiple sentences, and evaluate the output of the MLLM every $T$ sentences (concretely, we use the delimiter ``.'').
As $T$ grows, the reward model will evaluate longer and longer outputs.
As $T$ gets larger than the largest output length (equivalent behavior to $T=\infty)$, only complete outputs concluded with an \texttt{<EOS>} token are evaluated, and one complete output is selected among them: this strategy is usually referred to as \emph{rejection sampling} or \emph{best-of-$k$} in the literature~\cite{brown2024large,snell2024scaling}.
Figure~\ref{fig:mrgd_diagram} provides a summary of our method.\looseness-1

\section{Experiments}
\label{sec:experiments}

We evaluate our multimodal reward-guided decoding strategy in mitigating object hallucinations in long captions, and study the trade-offs between object precision and recall, and between visual grounding and test-time compute.\looseness-1

\subsection{Experimental setup}
\label{sec:experimental_setup}

\paragraph{Training data.}
We train our reward model for evaluating object hallucination on a mixture of publicly available multimodal preference datasets where responses without hallucinations are preferred over those with hallucinations: LLaVA-RLHF~\cite{sun2023aligning} (9.4k), RLHF-V~\cite{yu2024rlhf} (5.7k), and POVID~\cite{zhou2024aligning} (17k).
In addition, we repurpose SugarCrepe~\cite{hsieh2024sugarcrepe} (7.5k) as preference data\footnote{We use the instruction ``\texttt{Describe this image}''.}.
We use an 80/20\% train/validation split for each dataset. To handle varying dataset sizes, we sample each minibatch such that it has roughly the same amount of examples from each dataset.\looseness-1

\noindent\textbf{Implementation details.}
We initialize our object hallucination reward model's backbone from PaliGemma\footnote{\texttt{google/paligemma-3b-pt-224}}, train the linear regression head from scratch and finetune the backbone with LoRA~\cite{hulora}.
We use an effective minibatch size of 256, warm up the learning rate from 0 to $1e^{-3}$ during the first 5\% of the first epoch and decay it to zero with a cosine schedule. We train the reward model for a single epoch.
For the object recall reward model, we use the open-vocabulary object detector OWLv2\footnote{\texttt{google/owlv2-base-patch16-ensemble}}~\cite{minderer2024scaling}, the word embedding model Sentence-BERT\footnote{\texttt{sentence-transformers/all-mpnet-base-v2}}~\cite{reimers2019sentence}
and the POS tagger from the Natural Language Toolkit (NLTK). We set the object similarity threshold $\tau{=}0.5$.
We use LLaVA-1.5$_{\text{7B}}$\footnote{\texttt{llava-hf/llava-1.5-7b-hf}}~\cite{liu2024improved}, Llama-3.2-Vision$_{\text{11B}}$\footnote{\texttt{meta-llama/Llama-3.2-11B-Vision-Instruct}}~\cite{dubey2024llama} and SmolVLM-2$_{\text{2.2B}}$\footnote{\texttt{HuggingFaceTB/SmolVLM2-2.2B-Instruct}} as our base MLLMs. We caption images with the prompt ``{\small \texttt{Describe this image in detail}}'' for LLaVA-1.5 and SmolVLM-2, and ``{\small \texttt{Describe this image in a few sentences}}'' for Llama-3.2-Vision. For guided decoding, unless otherwise specified, we use a sampling temperature of $t{=}1.0$ for LLaVA-1.5 and $t{=}0.2$ for Llama-3.2-Vision \and SmolVLM-2.\looseness-1

\noindent\textbf{Evaluation setup.}
We evaluate our method on two standard object hallucination benchmarks, CHAIR~\cite{rohrbach2018object} (5k) and AMBER~\cite{wang2023llm} (1k), and report instance-level and sentence-level hallucination rates (the inverse of object precision), using the metrics used in respective benchmarks -- $\mathrm{C}_i$ and $\mathrm{CHAIR}$ for instance-level, and $\mathrm{C}_s$ and $\mathrm{Hal.}$ for sentence level. We also report object recall using the $\mathrm{Rec.}$ and $\mathrm{Cov.}$ (short for coverage) metrics, and caption length (denoted by $\mathrm{Len.}$) to ensure our method generates meaningful captions rather than degenerating into object-less outputs (more details in Appendix~\ref{sec:evaluation_metrics}).\looseness-1

\begin{table*}[t]
    \caption{Results on object hallucination benchmarks for LLaVA-1.5. MRGD with $k{=}30$ and $T{=}1$. C$_i$/CHAIR: instance-level hallucination rate, C$_s$/Hal.: sentence-level hallucination rate, Rec./Cov.: object recall/coverage, BS@$k$: beam search with $k$ beams, $*$: reported results from \cite{sarkar2024mitigating}, $\dagger$: results computed by us running the original code,
    ?: the decoding strategy used is unclear from the paper.
    }
    \centering
    \small
    \begin{tabular}{llccccccccc}
        \toprule
        \multirow{2}{*}{Model} & \multirow{2}{*}{Decoding} & \multicolumn{4}{c}{COCO} & \multicolumn{3}{c}{AMBER} \\
        \cmidrule{3-6} \cmidrule{7-9}
         & strategy & C$_i$ ($\downarrow$) & C$_s$ ($\downarrow$) & Rec. ($\uparrow$) & Len. & CHAIR ($\downarrow$) & Hal. ($\downarrow$) & Cov. ($\uparrow$) \\
        \midrule
        \textit{Baselines} \\
        \multirow{3}{*}{LLaVA-1.5$_{\text{7B}}$~\cite{liu2024improved}} & Greedy  & 15.05 & 48.94 & 81.30 & 90.12 & 7.6 & 31.8 & 49.3 \\
         & Greedy + Prompting & 13.50 & 44.00 & 80.38 & 92.98 & 6.7 & 29.1 & 49.4 \\
         & BS@30 & 15.68 & 55.00 & 81.62 & 101.89 & 11.2 & 41.4 & 46.1 \\
        \midrule
        \textit{Fine-tuning approaches} \\
        LLaVA-RLHF$_{\text{7B}}^\dagger$~\cite{sun2023aligning} & Greedy & 16.09 & 57.24 & 81.34 & 119.82 & 10.2 & 48.7 & 53.0 \\
        HA-DPO$^*$~\cite{zhao2023beyond} & BS@5  & 11.0 & 38.2 & - & 91.0 & 6.7 & 30.9 & 49.8 \\
        POVID~\cite{zhou2024aligning} & ? & 5.4 & 31.8 & - & - & - & - & - \\
        EOS$^*$~\cite{yue2024less} & Greedy  & 12.3 & 40.2 & - & 79.7 & 5.1 & 22.7 & 49.1 \\
        HALVA$_{7B}$~\cite{sarkar2024mitigating} & ?  & 11.7 & 41.4 & - & 92.2 & 6.6 & 32.2 & 53.0 \\
        CSR~\cite{zhou2024calibrated} & BS@5 & 7.3 & 28.0 & - & - & - & - & - \\
        \citet{liu2024investigating} & ? & 14.5 & 55.0 & 79.2 & 107.5 & 6.5 & 31.7 & 50.9 \\
        mDPO~\cite{wang2024mdpo} & ? & 9.8 & 35.7 & - & - & 4.4 & 24.5 & 52.4 \\
        \midrule
        \textit{Guided decoding approaches} \\
        \multirow{7}{*}{LLaVA-1.5$_{\text{7B}}$} & VCD$^\dagger$~\cite{leng2024mitigating} & 15.76 & 54.18 & 81.66 & 102.91 & 9.7 & 42.8 & 51.6 \\
         & CGD$^\dagger$~\cite{deng2024seeing} & 9.48 & 37.48 & 80.11 & 88.59 & 5.1 & 24.0 & 48.3 \\[0.5em]
         & MRGD$_{w=1.0}$ & \textbf{4.53} & \textbf{18.19} & 76.04 & 95.90 & 3.4 & \textbf{15.9} & 52.4 \\
         & MRGD$_{w=0.75}$ & 4.76 & 19.28 & 76.84 & 96.17 & \textbf{3.2} & 17.3 & 56.7 \\
         & MRGD$_{w=0.5}$ & 5.34 & 22.54 & 78.63 & 97.96 & 4.4 & 25.4 & 60.8 \\
         & MRGD$_{w=0.25}$ & 7.67 & 32.63 & 81.56 & 105.34 & 6.5 & 37.7 & 63.8 \\
         & MRGD$_{w=0.0}$ & 24.20 & 73.42 & \textbf{85.23} & 108.92 & 14.8 & 65.0 & \textbf{64.3} \\
        \bottomrule
    \end{tabular}
    \label{tab:results_llava}
    \vspace{-1em}
\end{table*}

\subsection{Reward model evaluation}

We first evaluate the performance of $r_\text{hal}$ on a held-out validation set from our preference data. We define accuracy as the fraction of times the reward model assigns higher scores to chosen vs. rejected responses, i.e. $r^\theta_{\text{hal}}(x_v, x_q, y^+) > r^\theta_{\text{hal}}(x_v, x_q, y^-)$. We obtain an average validation accuracy of $82.05\%$.
We also evaluate $r_\text{hal}$ on 5000 examples from VLFeedback~\cite{li2024vlfeedback} (not part of $r_\text{hal}$'s training data), selecting best and worst responses for each example, and obtain $67.68\%$ accuracy.
This is in line with the performance of well-behaved reward models~\cite{lambert2024rewardbench}.

We also evaluate $r_\text{rec}$'s object detector on COCO images and obtain a precision of 63.16\% and a recall of 55.83\%.
Similarly, we evaluate $r_\text{rec}$'s POS tagger on COCO captions and obtain a precision of 67.04\% and a recall of 54.54\%.
Although $r_\text{rec}$ is an imperfect estimator, we empirically verify it helps improve object recall.

\subsection{Comparison to baselines and existing methods}

We compare MRGD with existing hallucination mitigation methods based on fine-tuning~\cite{zhao2023beyond,zhou2024aligning,yue2024less,sarkar2024mitigating,zhou2024calibrated,liu2024investigating,wang2024mdpo}
and guided decoding~\cite{leng2024mitigating,deng2024seeing} for the LLaVA-1.5 base model. Existing methods were selected based on recency, comparability and code/checkpoint availability (more details in Appendix~\ref{sec:reporting}).
We also implement a prompting baseline based on requesting the desired response properties in the input prompt (more details in Appendix~\ref{sec:more_baselines}).
For MRGD, we choose the best performing variant w.r.t. the number of samples $k$, the reward evaluation period $T$, and the temperature $t$.

\begin{table*}[t]
    \caption{Results on object hallucination benchmarks for Llama-3.2-Vision and SmolVLM-2. MRGD with $k{=}30$ and $T{=}1$. C$_i$/CHAIR: instance-level hallucination rate, C$_s$/Hal.: sentence-level hallucination rate, Rec./Cov.: object recall/coverage.}
    \centering
    \small
    \begin{tabular}{llccccccccc}
        \toprule
        \multirow{2}{*}{Model} & \multirow{2}{*}{Decoding} & \multicolumn{4}{c}{COCO} & \multicolumn{3}{c}{AMBER} \\
        \cmidrule{3-6} \cmidrule{7-9}
         & strategy & C$_i$ ($\downarrow$) & C$_s$ ($\downarrow$) & Rec. ($\uparrow$) & Len. & CHAIR ($\downarrow$) & Hal. ($\downarrow$) & Cov. ($\uparrow$) \\
        \midrule
        \multirow{5}{*}{Llama-3.2-Vision$_{\text{11B}}$~\cite{dubey2024llama}} & Greedy & 5.82 & 20.52 & 71.45 & 91.61 & 6.1 & 35.4 & 66.0 \\
         & Greedy + Prompting & 6.14 & 25.24 & 71.92 & 160.10 & 5.5 & 39.4 & 65.1 \\[0.5em]
         & MRGD$_{w=1.0}$ & \textbf{4.38} & \textbf{15.50} & 69.54 & 87.18 & \textbf{4.1} & \textbf{25.2} & 65.0 \\
         & MRGD$_{w=0.5}$ & 4.76 & 16.75 & 71.47 & 87.53 & 4.8 & 28.8 & 68.9 \\
         & MRGD$_{w=0.0}$ & 6.89 & 24.50 & \textbf{74.13} & 92.33 & 7.2 & 42.2 & \textbf{70.7} \\
        \midrule
        \multirow{4}{*}{SmolVLM-2$_{\text{2.2B}}$~\cite{marafioti2025smolvlm}} & Greedy & 6.06 & 20.08 & 70.17 & 85.92 & 5.3 & 25.7 & 58.4 \\[0.5em]
         & MRGD$_{w=1.0}$ & \textbf{4.32} & \textbf{14.38} & 68.04 & 79.27 & \textbf{3.8} & \textbf{18.2} & 58.7 \\
         & MRGD$_{w=0.5}$ & 4.98 & 16.46 & 69.93 & 78.47 & 4.4 &21.9 & 62.5 \\
         & MRGD$_{w=0.0}$ & 7.09 & 25.00 & \textbf{73.22} & 84.39 & 6.6 & 33.4 & \textbf{64.1} \\
        \bottomrule
    \end{tabular}
    \label{tab:results_llamav_smolvlm}
\end{table*}

\begin{figure*}[ht]
    \hspace*{-0.25cm}
    \begin{subfigure}[b]{0.48\textwidth}
        \centering
        \includegraphics[width=\linewidth]{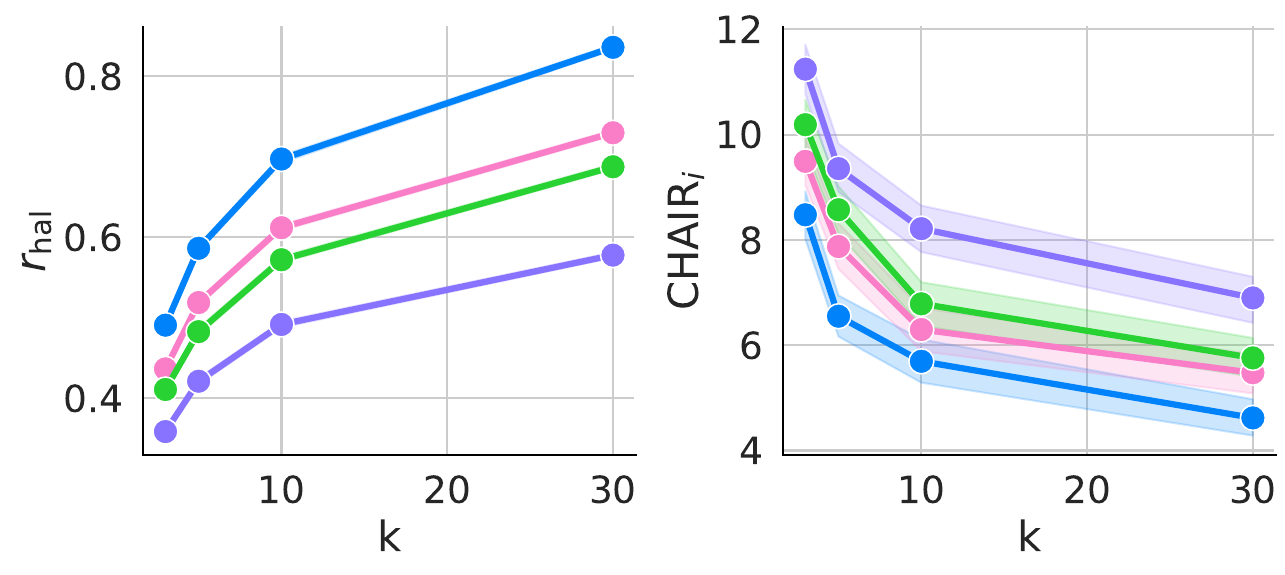}
        \caption{Reward value $r_\text{hal}$ (left) and CHAIR$_i$ (right), MRGD with $w{=}1.0$.}
        \label{fig:hal_vs_k}
    \end{subfigure}
    \begin{subfigure}[b]{0.54\textwidth}
        \centering
        \includegraphics[width=\linewidth]{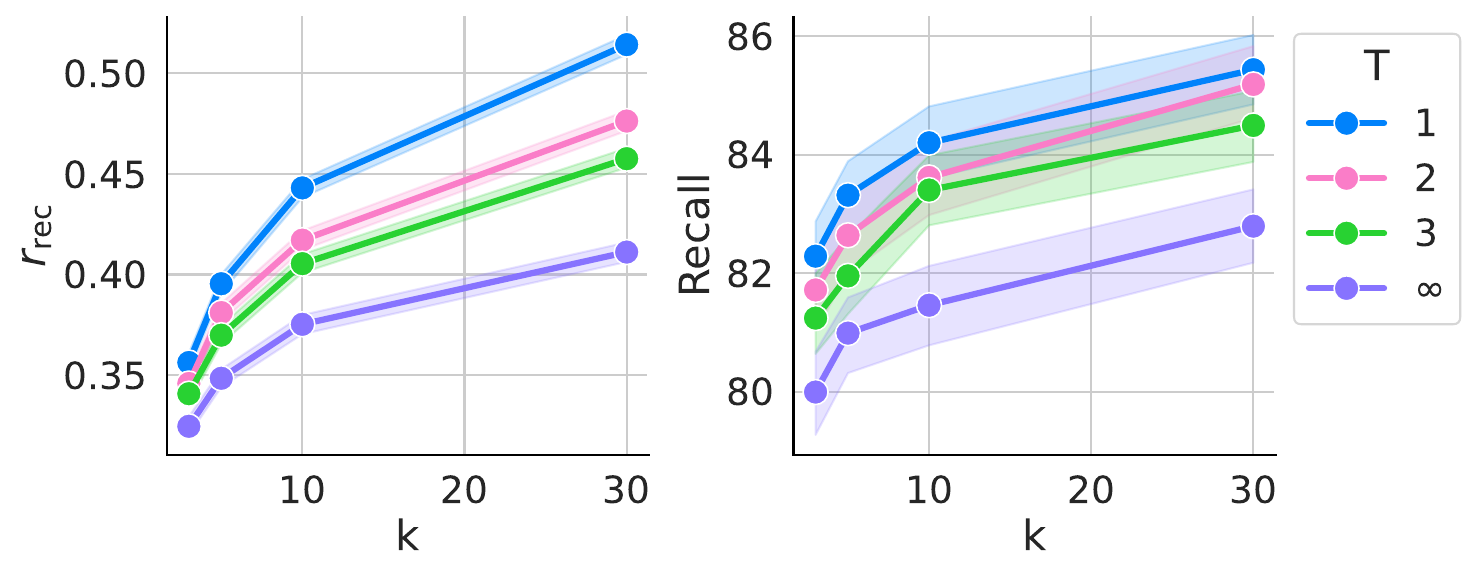}
        \caption{Reward value $r_\text{rec}$ (left) and Recall (right), MRGD with $w{=}0.0$.}
        \label{fig:rec_vs_k}
    \end{subfigure}
    \caption{LLaVA-1.5 on COCO. Leveraging the reward model to guide the generation more often (lower $T$) improves sample-efficiency.}
    \label{fig:reward_vs_k}
    \vspace{-1em}
\end{figure*}

Table~\ref{tab:results_llava} shows MRGD with $w{=}1$ considerably reduces object hallucinations w.r.t. greedy decoding, at the expense of a moderate decrease in object recall/coverage. For instance, on the COCO benchmark, CHAIR$_i$ is reduced by $\sim$70\% (from 15.05\% with greedy decoding to 4.53\% with MRGD) while recall is only reduced by 6.5\%. By combining both reward models with $w{=}0.5$, recall is substantially increased w.r.t. $w{=}1.0$ (2.6\% on COCO and 8.4\% on AMBER), without overly increasing the hallucination rate (0.8\% on COCO and 1\% AMBER). When $w{=}0$, MRGD achieves state-of-the-art results on object recall/coverage at the cost of a higher hallucination rate (e.g., CHAIR$_i$ is increased by 60.8\% w.r.t. greedy decoding).

We also observe the optimal operating point $w^*$---mitigating object hallucinations without loosing recall---varies by benchmark, with $w^*{\approx}0.25$ for COCO and $w^* = 1.0$ for AMBER.
An analysis of 500 images from COCO and AMBER reveals that COCO images have an average of 21.4 detected objects, compared to 9.9 for AMBER, resulting in systematically lower $r_\text{rec}$ values for COCO. Therefore, the optimal $w$ assigns more weight to $r_\text{rec}$ (lower $w$) for COCO than for AMBER.

Compared to prior visual hallucination mitigation methods, MRGD consistently surpasses the performance of methods which fine-tune the base MLLM, while offering greater flexibility and more granular control over the MLLM's behavior.
For guided decoding approaches, we see MRGD also outperforms
CGD~\cite{deng2024seeing}, achieving $\sim$50\% lower hallucination rate on COCO and $\sim$30\% on AMBER.
Surprisingly, LLaVA-RLHF and VCD~\cite{leng2024mitigating} exhibit a higher hallucination rate than greedy decoding on captioning hallucination benchmarks, which were not considered in the original papers;
instead, they limited their evaluation to discriminative hallucination benchmarks consisting of visual questions.
This suggests that generative (captioning) and discriminative (VQA) hallucination benchmarks may not be as strongly correlated as previously assumed.\looseness-1

Lastly, Table~\ref{tab:results_llamav_smolvlm} demonstrates that MRGD continues to be effective when applied to newer and architecturally diverse MLLMs such as Llama-3.2-Vision and SmolVLM-2. Notably, our reward models can be readily applied to new MLLMs without retraining.\looseness-1

\subsection{Applying MRGD on top of RLHF}

While the hallucination mitigation literature focuses primarily on the instruction fine-tuned LLaVA-1.5 model, here we assess the effectiveness of our method with more recent MLLMs that have been already fine-tuned with RLHF. We apply MRGD on top of Llama-3.2-Vision, which has undergone a preference alignment phase (with DPO~\cite{rafailov2024direct}) after instruction fine-tuning.
Crucially, its multimodal preference data includes visual grounding examples~\cite{dubey2024llama}, which makes it less prone to hallucinations.\looseness-1

Table~\ref{tab:results_llamav_smolvlm} shows that MRGD is also effective when applied to Llama-3.2-Vision. As expected, the improvement in object precision and recall is smaller compared to LLaVA-1.5 since Llama-3.2-Vision already starts from a better level of visual grounding. However, MRGD can further mitigate object hallucinations: when guiding decoding with the reward model for object hallucinations ($w{=}1.0$), we observe a $\sim$1\% and $\sim$2\% reduction in instance-level hallucinations for COCO and AMBER respectively, a $\sim$5\% and $\sim$10\% decrease in sentence-level hallucinations, and only a slight decrease in object recall.\looseness-1

One interesting observation is that Llama-3.2-Vision's object recall on COCO is considerably lower ($\sim$-10\%) than that of LLaVA-1.5, which may be due to more conservative outputs as a consequence of preference and safety fine-tuning. Guiding decoding with the reward model for object recall ($w{=}0.0$) boosts object recall in the generated captions by $\sim$2.7\% on COCO and $\sim$4.7\% on AMBER.

\begin{table*}[ht]
    \centering
    \small
    \caption{Ablation results for LLaVA-1.5$_\text{7B}$. MRGD with $k{=}30$, $T{=}1$, $w{=}0.5$. MRGD$_{\text{PG2}}$: using PaliGemma-2 instead of PaliGemma for $r_\text{hal}$, MRGD$_{+\text{RLAIF-V}}$: removing RLAIF-V from the original data mix for $r_\text{hal}$, MRGD$_{+\text{RLAIF-V}-\text{POVID}}$: adding RLAIF-V and removing POVID from the original data mix, MRGD$_{\text{DETR}}$: using DETR instead of OWLv2 as object detector for $r_\text{rec}$, and MRGD$_{\tau{=}x}$: using $x$ instead of $0.5$ as semantic similarity threshold for $r_\text{rec}$.\looseness-1}
    \begin{tabular}{lccccccc}
        \toprule
        \multirow{2}{*}{Decoding strategy} & \multicolumn{4}{c}{COCO} & \multicolumn{3}{c}{AMBER} \\
        \cmidrule{2-5} \cmidrule{6-8}
         & C$_i$ ($\downarrow$) & C$_s$ ($\downarrow$) & Rec. ($\uparrow$) & Len. & CHAIR ($\downarrow$) & Hal. ($\downarrow$) & Cov. ($\uparrow$) \\
        \midrule
        Greedy & 15.05 & 48.94 & 81.30 & 90.12 & 7.6 & 31.8 & 49.3 \\
        MRGD & 5.34 & 22.54 & 78.63 & 97.96 & 4.4 & 25.4 & 60.8 \\
        \midrule
        \textit{$r_\text{hal}$ variants} \\
        MRGD$_{\text{PG2}}$ & 5.88 & 27.07 & 78.76 & 105.25 & 4.1 & 25.0 & 59.6 \\
        MRGD$_{+\text{RLAIF-V}}$ & 7.83 & 29.68 & 77.54 & 94.26 & 6.3 & 33.2 & 57.1 \\
        MRGD$_{+\text{RLAIF-V}-\text{POVID}}$ & 8.17 & 34.08 & 79.03 & 104.04 & 5.1 & 29.3 & 59.9 \\
        \midrule
        \textit{$r_\text{rec}$ variants} \\
        MRGD$_{\text{DETR}}$ & 5.37 & 23.76 & 82.04 & 99.24 & 4.0 & 19.8 & 53.5 \\
        MRGD$_{\tau{=}0.2}$ & 5.89 & 24.46 & 78.09 & 106.86 & 4.3 & 22.8 & 54.5 \\
        MRGD$_{\tau{=}0.9}$ & 5.00 & 20.96 & 78.36 & 98.09 & 4.0 & 22.3 & 61.2 \\
        \bottomrule
    \end{tabular}
    \label{tab:rm_ablations}
\end{table*}

\subsection{Visual grounding vs. compute trade-off}

By varying the number of samples $k$ and the evaluation period $T$, we can control the trade-off between the degree of visual grounding in the generated outputs and the amount of compute used during decoding, for fixed guidance strength $w$ and temperature $t$: expanding the search space (increasing $k$) and evaluating more frequently (decreasing $T$) increases visual grounding but also the required compute.\looseness-1

\begin{figure}[t]
    \centering
    \includegraphics[width=\linewidth]{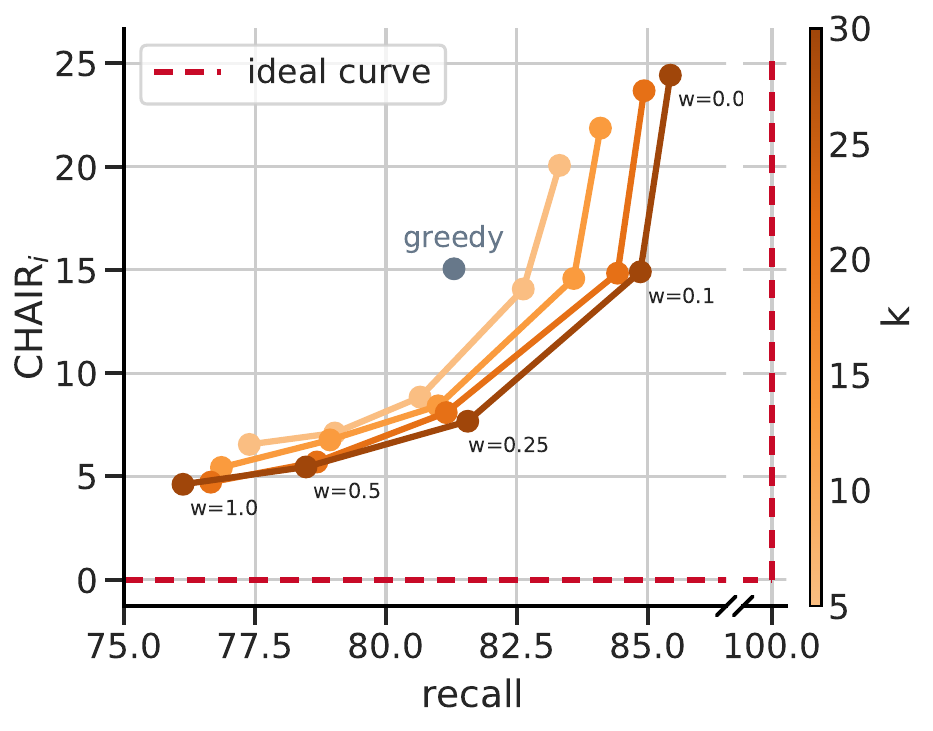}
    \caption{Object precision and recall for LLaVA-1.5 on COCO, with $T{=}1$.
    Varying $w$ modulates the precision-recall trade-off under a fixed compute budget, while increasing compute via a larger $k$ improves both precision and recall, significantly surpassing greedy search and bringing the trade-off curve closer to the ideal.\looseness-1}
    \label{fig:chair_vs_recall}
\end{figure}

Figure~\ref{fig:hal_vs_k} shows how the reward value $r_\text{hal}$ and hallucination rate (CHAIR$_i$) evolve as we increase the number of samples $k{=}\{3, 5, 10, 30\}$ and as we vary the evaluation period $T$. As expected, we observe the hallucination rate decreases as we increase $k$. Notably, MRGD (with $T{=}1$) is considerably more compute-efficient than naive rejection sampling (equivalent to $T{=}\infty$).
For example,
MRGD with $k{=}5$ achieves a lower hallucination rate than rejection sampling with $k{=}30$, making MRGD over $6\times$ more sample-efficient than rejection sampling.
We observe similar trends for object recall in Figure~\ref{fig:rec_vs_k}.
Note that evaluating more frequently also increases computational cost, but to a much lesser extent than generating, since the reward models are considerably smaller than the base MLLM and evaluation only requires a single forward pass.\looseness-1

\subsection{Object precision vs. recall trade-off}

Figure~\ref{fig:chair_vs_recall} shows the trade-off between hallucination rate (CHAIR$_i$) and object recall when varying the guidance strength $w{=}\{0.0, 0.1, 0.25, 0.5, 1.0\}$ for fixed $k$ and $T$. We observe a lower $w$ leads to higher recall and  lower precision (higher CHAIR$_i$) and vice-versa.
And the trade-off curve gets closer to the ideal curve with higher $k$.
This suggests that there is an inherent trade-off between precision and recall in MLLMs.
However, our approach gives a user the flexibility to choose the operating point (by choosing a value for $w$ and $k$) that suits their needs at inference time.\looseness-1

\subsection{Preference data mix for $r_\text{hal}$}

To understand the impact of different preference data compositions on the quality of $r_\text{hal}$, we conduct an ablation over the datasets used for its training. Our base reward model is trained on a mixture of LLaVA-RLHF~\cite{sun2023aligning} (9.4k), RLHF-V~\cite{yu2024rlhf} (5.7k), and POVID~\cite{zhou2024aligning} (17k). We consider an additional preference dataset, RLAIF-V~\cite{yu2024rlaif} (83k), which contains 2.6$\times$ more examples than all previous datasets combined. We train two additional variants: (1) adding RLAIF-V and (2) adding RLAIF-V while removing POVID. As shown in Table~\ref{tab:rm_ablations}, both adding RLAIF-V and removing POVID lead to notable performance degradation, highlighting the importance of carefully choosing the preference data mix to train $r_\text{hal}$.\looseness-1

\begin{figure*}[ht]
    \vspace{-1em}
    \begin{subfigure}[b]{\textwidth}
        \centering
        \includegraphics[width=0.95\textwidth]{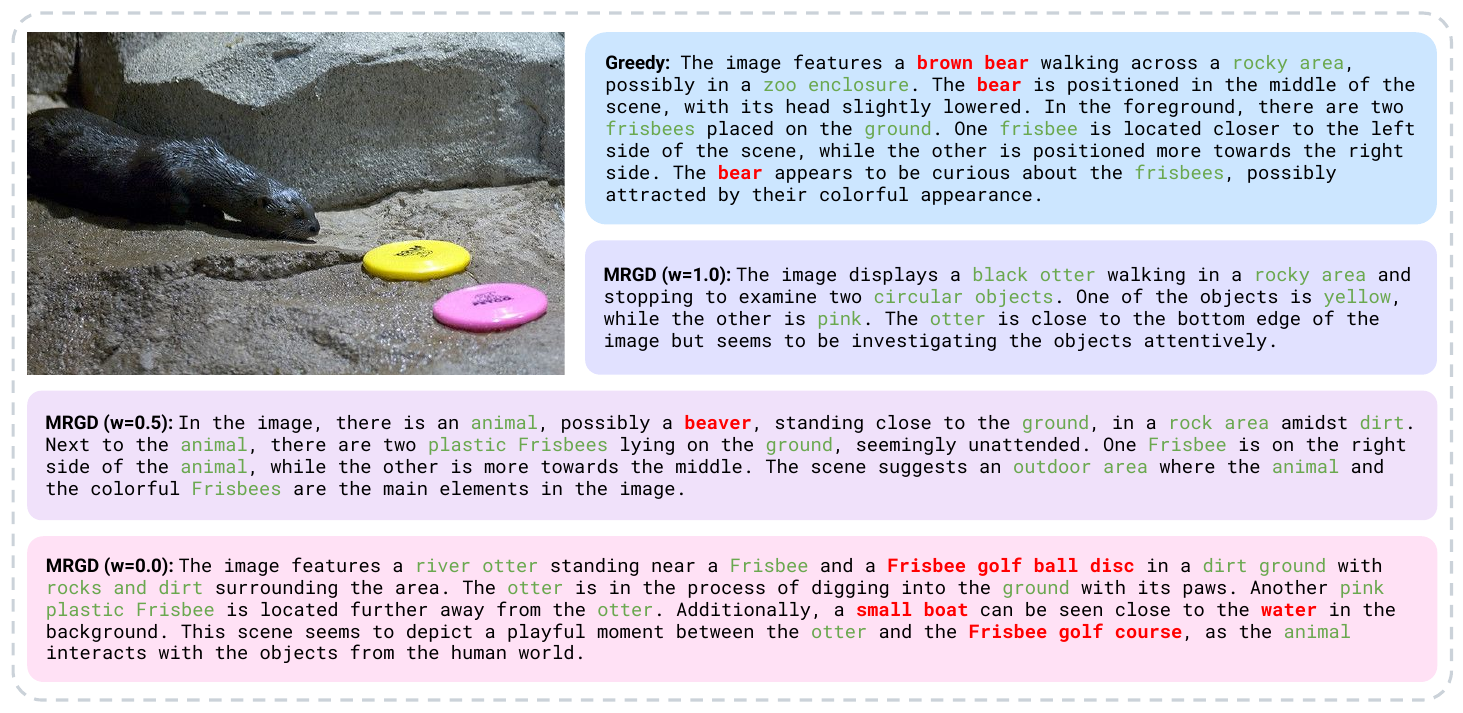}
        \caption{}
        \label{fig:qualitative1}
    \end{subfigure}
    \\
    \begin{subfigure}[b]{\textwidth}
        \centering
        \includegraphics[width=0.95\textwidth]{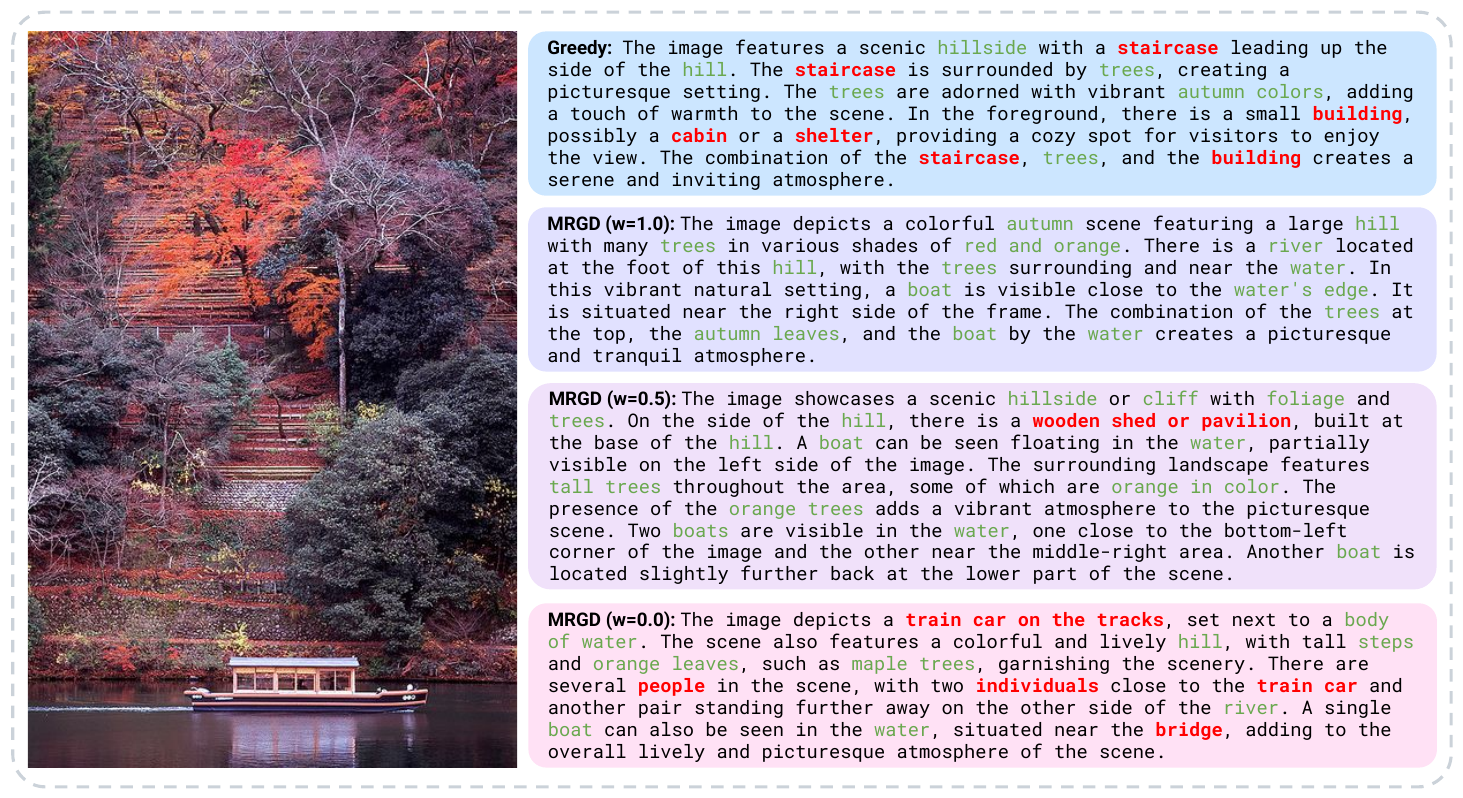}
        \caption{}
        \label{fig:qualitative2}
    \end{subfigure}
    \vspace{-1.5em}
    \caption{Selected qualitative examples for LLaVA-1.5 using the default greedy decoding and our proposed MRGD strategy with different $w$'s ($k{=}30, T{=}1$). Correct objects are highlighted in {\color{YellowGreen} green} and hallucinated objects in {\color{Red} red}. New lines in captions are omitted for brevity.}
    \label{fig:qualitative}
    \vspace{-1em}
\end{figure*}

\subsection{MRGD's robustness to reward models' quality}

To further assess the robustness of MRGD to variations in reward model quality, we evaluate the performance of our approach for a variant of $r_\text{hal}$ with a different model backbone (PaliGemma-2$_\text{3B}$\footnote{\texttt{google/paligemma2-3b-pt-224}}~\cite{steiner2024paligemma} instead of PaliGemma), and several variants of $r_{rec}$: with different object detector (DETR~\cite{carion2020end} instead of OWLv2) and different semantic similarity thresholds ($\tau$).
In Table~\ref{tab:rm_ablations}, we observe that
(1) upgrading $r_\text{hal}$'s backbone yields similar performance, 
(2) using DETR for $r_\text{rec}$ performs similarly for COCO and decreases both hallucinations and coverage for AMBER, 
and (3) MRGD remains effective when varying $\tau$.
Overall, all ablated variants significantly outperform the greedy search baseline on most metrics, demonstrating the effectiveness and robustness of our approach across reward design choices.\looseness-1

\begin{table}[t]
    \caption{Average decoding times on a NVIDIA A100 GPU.}
    \centering
    \small
    \begin{tabular}{lcc}
        \toprule
        Decoding strategy & C$_i$ ($\downarrow$) & Time (s) ($\downarrow$) \\
        \midrule
        Greedy (vLLM) & 15.05 & $1.35 \pm 0.49 $ \\
        BS@10 (HuggingFace) & 15.80 & $5.26 \pm 1.15 $ \\
        BS@30 (HuggingFace) & 15.68 & $15.14 \pm 2.75 $ \\
        MRGD@10 (vLLM) & 5.76 & $6.79 \pm 1.93 $ \\
        MRGD@30 (vLLM) & 4.53 & $14.78 \pm 4.38 $ \\
        \bottomrule
    \end{tabular}
    \label{tab:latency}
\end{table}

\subsection{Computational cost and latency}

In terms of computational cost, finetuning PaliGemma on 30.6k preference examples requires only $\sim$9 minutes on 8$\times$ NVIDIA H100 GPUs. Furthermore, as shown in Table~\ref{tab:latency}, spending more test-time compute with MRGD helps reduce hallucinations. Notably, while MRGD with $k{=}30$ generates $\sim$30$\times$ more text w.r.t. greedy decoding, its latency increase is significantly less than 30$\times$ due to batched generation.\looseness-1

\subsection{Qualitative analysis}

Figure~\ref{fig:qualitative} illustrates qualitative differences in generated captions between using the default greedy decoding and our MRGD strategy, for the same input images. Greedy decoding produces captions which contain either more or less objects than those actually present in the images. For instance, in
Figure~\ref{fig:qualitative1}, the greedy caption incorrectly identifies the animal as a ``brown bear'', significantly hurting object precision, though it does correctly describe the ``colorful frisbees'' and ``rocky terrain''. When using MRGD with $w{=}1.0$, the model produces an accurate caption with no hallucinations, correctly naming the ``black otter'' and both ``circular objects'' while capturing the investigative behavior, though it omits finer scene details like the ground texture. With $w{=}0.5$, the model misidentifies the animal as a ``beaver'' but achieves a good object recall, referencing the ``frisbees'', their positions, and the ``dirt'' and ``rock area''. At $w{=}0.0$, the caption achieves the highest object recall, correctly mentioning the ``otter'', ``frisbees'', and the surrounding environment with ``rocks'' and ``dirt'', but introduces multiple hallucinated elements such as a ``small boat close to the water'' and a ``disc golf ball'', which reduce object precision.
Similarly, in
Figure~\ref{fig:qualitative2}, the greedy caption includes a hallucinated ``staircase'' and a ``building'', while failing to mention the boat or river. With MRGD and $w{=}1.0$, the caption correctly identifies the ``hill'', ``trees``, ``river``, ``autumn leaves'', and ``boat'' without hallucinations, but it omits the terraced steps. At $w{=}0.5$, the caption recalls the ``hillside or cliff'', ``foliage``, ``trees'', ``water'' and a ``boat'', but introduces false objects like a ``wooden shed or pavilion'' and additional boats. At $w{=}0.0$, the caption reaches the highest object recall by naming the ``hill'', ``steps'', ``leaves'', ``body of water'', and a ``boat'', but suffers from severe hallucinations, adding a ``train car'', ``tracks'', ``people'', and a ``bridge''.
\looseness-1

\section{Conclusion}
\label{sec:conclusion}

In this paper, we presented MRGD, a reward-guided decoding method for MLLMs based on multimodal reward models for visual grounding. We built two reward models---one evaluating object precision in captions, another evaluating object recall---used in an iterative search process that evaluates candidate responses against their combined reward values.
Our methodology enables on-the-fly controllability of MLLM-generated captions along two axes: controlling the object precision/recall trade-off by adjusting the weight of each reward model, and balancing test-time compute vs. visual grounding by varying the search breadth and frequency. Our method provides significant controllability over MLLM inference while matching or surpassing existing hallucination mitigation methods.\looseness-1

\paragraph{Limitations and future work.}
In this work, we focused on mitigating \emph{object} hallucinations primarily due to their ease of automatic evaluation, but other important visual hallucinations exist -- related to attributes, count, spatial relationships, negation, and more -- which we leave for future work.
In addition, we would like to continue exploring: (1) building reward models for semantically incomplete outputs, (2) extending MRGD to discriminative hallucination tasks (e.g., POPE~\cite{li2023evaluating}), and (3) gradient-based optimization instead of search-based approaches.

\section*{Acknowledments}

We thank Saba Ahmadi, Qian Yang and Shravan Nayak for providing valuable feedback on an earlier draft of this work. During this project, Aishwarya Agrawal was supported by the Canada CIFAR AI Chair award.

{
    \small
    \bibliographystyle{template/ieeenat_fullname}
    \bibliography{main}
}

\clearpage
\setcounter{page}{1}
\maketitlesupplementary

\section{Experiments}

\subsection{Details on evaluation metrics}
\label{sec:evaluation_metrics}

We evaluate object precision and recall with standard metrics from the corresponding benchmarks, defined as follows.\looseness-1

\paragraph{$\boldsymbol{\mathrm{CHAIR}_i}$ ($\boldsymbol{\mathrm{C}_i}$)~\citep{rohrbach2018object}, $\boldsymbol{\mathrm{CHAIR}}$~\citep{wang2023llm}.}
Measure the fraction of hallucinated objects in the generated captions.

\begin{equation*}
    \mathrm{C}_i/\mathrm{CHAIR} = \frac{|\{\text{hallucinated objects}\}|}{|\{\text{all mentioned objects}\}|}
\end{equation*}

\paragraph{$\boldsymbol{\mathrm{CHAIR}_s}$ ($\boldsymbol{\mathrm{C}_s}$)~\citep{rohrbach2018object}, $\boldsymbol{\mathrm{Hal.}}$~\cite{wang2023llm}.}
Measure what fraction of generated captions include a hallucinated object.

\begin{equation*}
    \mathrm{C}_s/\mathrm{Hal.} = \frac{|\{\text{captions with a hallucinated object}\}|}{|\{\text{all captions}\}|}
\end{equation*}

\paragraph{Recall ($\boldsymbol{\mathrm{Rec.}}$), Coverage ($\boldsymbol{\mathrm{Cov.}}$)~\cite{wang2023llm}.}
Measure the fraction of ground-truth objects covered in the generated captions.

\begin{equation*}
    \mathrm{Rec.}/\mathrm{Cov.} = \frac{|\{\text{correct objects}\}|}{|\{\text{all ground-truth objects}\}|}
\end{equation*}

\subsection{Details on reporting results of existing methods}
\label{sec:reporting}

In Table~\ref{tab:results_llava}, we report results for existing hallucination mitigation methods from the best source available. Unless otherwise specified, values are directly copied from the corresponding papers. For HA-DPO~\cite{zhao2023beyond} and EOS~\cite{yue2024less}, values are copied from \citet{sarkar2024mitigating} since their evaluation setup matches ours. For LLaVA-RLHF~\cite{sun2023aligning} and VCD~\cite{leng2024mitigating}, we compute results by generating captions with the original code and evaluating them on CHAIR~\cite{rohrbach2018object} and AMBER~\cite{wang2023llm}, since the original papers do not report hallucination results on these benchmarks.
For CGD~\cite{deng2024seeing}, we also run the original code to generate captions for both AMBER and the full standard set of 5000 examples in the CHAIR benchmark (instead of the 500-example subset used by \citet{deng2024seeing}).

\subsection{Prompting baseline}
\label{sec:more_baselines}

We propose multimodal reward-guided decoding (MRGD) as a method to control the behavior of MLLMs at inference time. A common approach to steer the behavior of LLMs at inference time is prompting~\cite{brown2020language}. Here, we apply the same idea to MLLMs as an alternative approach to control their behavior. To mitigate visual hallucinations in image captioning, we use the instruction
``{\small \texttt{\{captioning instruction\}. Provide an accurate and objective description, focusing on verifiable visual elements such as colors, textures, shapes, and compositions. Avoid making assumptions, inferences, or introducing information not present in the image}}'',
where the captioning instruction is the one described in Section~\ref{sec:experimental_setup}: ``{\small \texttt{Describe this image in detail}}'' for LLaVA-1.5 and ``{\small \texttt{Describe this image in a few sentences}}'' for Llama-3.2-Vision. We maintain greedy decoding for the prompting baselines.
In Tables \ref{tab:results_llava} and \ref{tab:results_llamav_smolvlm}, we observe that prompting slightly reduces object hallucinations compared to greedy decoding for LLaVA-1.5, while for Llama-3.2-Vision, surprisingly, it does not help much and, in fact, it increases the sentence-level hallucination rate (CHAIR$_s$ and Hal.). Instead, with LLaVA-1.5 on COCO, for a similar level of object recall ($\sim$81\%), MRGD with $w{=}0.25$ achieves better object precision by $\sim$5.8\% CHAIR$_i$ and $\sim$11.4\% CHAIR$_s$ compared to prompting. This suggests that prompting is not a very effective strategy to steer MLLMs towards complex behaviors such as reducing visual hallucinations.

\begin{table*}[ht]
    \caption{Additional results for LLaVA-1.5$_\text{7B}$. MRGD with $k{=}30$ and $T{=}1$. MRGD$_{\text{PaliGemma}}$: MRGD using PaliGemma fine-tuned on preference data for $r_\text{hal}$, MRGD$_{\text{SigLIP}}$: MRGD using off-the-shelf SigLIP for $r_\text{hal}$.
    }
    \centering
    \small
    \begin{tabular}{lccccccccc}
        \toprule
        \multirow{2}{*}{Decoding strategy} & \multicolumn{4}{c}{COCO} & \multicolumn{3}{c}{AMBER} \\
        \cmidrule{2-5} \cmidrule{6-8}
         & C$_i$ ($\downarrow$) & C$_s$ ($\downarrow$) & Rec. ($\uparrow$) & Len. & CHAIR ($\downarrow$) & Hal. ($\downarrow$) & Cov. ($\uparrow$) \\
        \midrule
        Greedy & 15.05 & 48.94 & 81.30 & 90.12 & 7.6 & 31.8 & 49.3 \\
        MRGD$_{\text{PaliGemma},w=1.0}$ & 4.53 & 18.19 & 76.04 & 95.90 & 3.4 & 15.9 & 52.4 \\
        MRGD$_{\text{PaliGemma},w=0.75}$ & 4.76 & 19.28 & 76.84 & 96.17 & 3.2 & 17.3 & 56.7 \\
        MRGD$_{\text{PaliGemma},w=0.5}$ & 5.34 & 22.54 & 78.63 & 97.96 & 4.4 & 25.4 & 60.8 \\
        MRGD$_{\text{PaliGemma},w=0.25}$ & 7.67 & 32.63 & 81.56 & 105.34 & 6.5 & 37.7 & 63.8 \\
        MRGD$_{w=0.0}$ & 24.20 & 73.42 & 85.23 & 108.92 & 14.8 & 65.0 & 64.3 \\
        \midrule
        MRGD$_{\text{SigLIP},w=1.0}$ & 7.19 & 28.00 & 73.71 & 92.73 & 6.0 & 30.1 & 48.5 \\
        MRGD$_{\text{SigLIP},w=0.75}$ & 7.57 & 29.58 & 74.30 & 93.17 & 6.1 & 30.3 & 50.0 \\
        MRGD$_{\text{SigLIP},w=0.5}$ & 8.17 & 32.88 & 75.96 & 94.93 & 6.3 & 33.3 & 53.4 \\
        MRGD$_{\text{SigLIP},w=0.25}$ & 10.84 & 43.58 & 79.50 & 99.57 & 8.5 & 46.2 & 57.8 \\
        \bottomrule
    \end{tabular}
    \label{tab:more_results}
\end{table*}

\subsection{Using SigLIP for $r_\text{hal}$}

CGD~\cite{deng2024seeing} can be viewed as a particular instance of MRGD when using off-the-shelf SigLIP as the reward model for object hallucinations and removing the combination of multiple reward models (i.e., setting $w{=}1.0$). Therefore, we also conduct an ablation of MRGD replacing PaliGemma fine-tuned on preference data (Section~\ref{sec:paligemma_preference}) with off-the-shelf SigLIP-SoViT-400m
\footnote{\texttt{google/siglip-so400m-patch14-384}}.
Due to SigLIP's limited context length of 64 tokens, we only evaluate the last generated sentence, unlike PaliGemma which receives the full prefix response (which may contain several sentences).
To ensure that the scores from multiple reward models are comparable and can be combined effectively, we normalize their ranges. In particular, since the effective range of SigLIP scores is much narrower than that of the reward model for object recall ($r_\text{rec} \in [0,1]$), we linearly rescale SigLIP scores $r \in \mathbb{R}^k$ to cover the range $[0,1]$: $r = (r - \min(r)) / (\max(r) - \min(r) + \epsilon)$, where $\min$ and $\max$ are computed across the set of candidate samples $Y$, and $\epsilon$ is a small value to avoid division by zero (in case all candidates obtained the same score).
In Table~\ref{tab:more_results}, we observe that when using a SigLIP-based $r_\text{hal}$, our MRGD strategy is still effective in reducing object hallucinations and enabling the user to trade off object precision and recall on-the-fly at inference time. However, SigLIP does not allow to reach the same level of object precision, and the trade-off with object recall is also worse. For instance, when $w{=}1.0$, MRGD$_{\text{PaliGemma}}$ achieves better object precision by $\sim$2.7\% CHAIR$_i$ and $\sim$9.8\% CHAIR$_s$, and better Recall by $\sim$2.3\% compared to MRGD$_{\text{SigLIP}}$.

\end{document}